\documentclass[10pt,twocolumn,letterpaper]{article}

\usepackage[pagenumbers]{wacv} 

\usepackage{graphicx}
\usepackage{amsmath}
\usepackage{amssymb}
\usepackage{booktabs}

\usepackage{enumitem}
\usepackage{algorithm}
\usepackage{algpseudocode}
\usepackage{amsmath}
\usepackage{amssymb}
\usepackage{float}
\usepackage{stfloats}
\usepackage[pagebackref,breaklinks,colorlinks]{hyperref}

\newcommand\blfootnote[1]{%
  \begingroup
  \renewcommand\thefootnote{}\footnote{#1}%
  \addtocounter{footnote}{-1}%
  \endgroup
}

\usepackage[capitalize]{cleveref}
\crefname{section}{Sec.}{Secs.}
\Crefname{section}{Section}{Sections}
\Crefname{table}{Table}{Tables}
\crefname{table}{Tab.}{Tabs.}

\def\wacvPaperID{1309} 
\def\confName{WACV}
\def\confYear{2024}

\begin{document}

\title{CAD - Contextual Multi-modal Alignment for Dynamic AVQA}

\author{Asmar Nadeem\(^1\), Adrian Hilton\(^1\), Robert Dawes\(^2\), Graham Thomas\(^2\), Armin Mustafa\(^1\)\\ 
\(^1\)Centre for Vision, Speech and Signal Processing (CVSSP), University of Surrey, United Kingdom. \\ \(^2\)BBC Research and Development,
United Kingdom.
\and
}
\maketitle

\begin{abstract}
In the context of Audio Visual Question Answering (AVQA) tasks, the audio visual modalities could be learnt on three levels: 1) Spatial, 2) Temporal, and 3) Semantic. Existing AVQA methods suffer from two major shortcomings; the audio-visual (AV) information passing through the network isn't aligned on Spatial and Temporal levels; and, inter-modal (audio and visual) Semantic information is often not balanced within a context; this results in poor performance. 
In this paper, we propose a novel end-to-end Contextual Multi-modal Alignment (CAD) network that addresses the challenges in AVQA methods by i) introducing a parameter-free stochastic Contextual block that ensures robust audio and visual alignment on the Spatial level; ii) proposing a pre-training technique for dynamic audio and visual alignment on Temporal level in a self-supervised setting, and iii) introducing a cross-attention mechanism to balance audio and visual information on Semantic level. 
The proposed novel CAD network improves the overall performance over the state-of-the-art methods on average by $9.4\%$ on the MUSIC-AVQA dataset. We also demonstrate that our proposed contributions to AVQA can be added to the existing methods to improve their performance without additional complexity requirements.
\end{abstract}

\blfootnote{Correspondence to: Asmar Nadeem
\( \mathnormal{<} \)asmar.nadeem@surrey.ac.uk\( \mathnormal{>} \).}


\section{Introduction}
\label{sec:intro}
\begin{figure}[t]
  \centering
   \includegraphics[width=\linewidth]{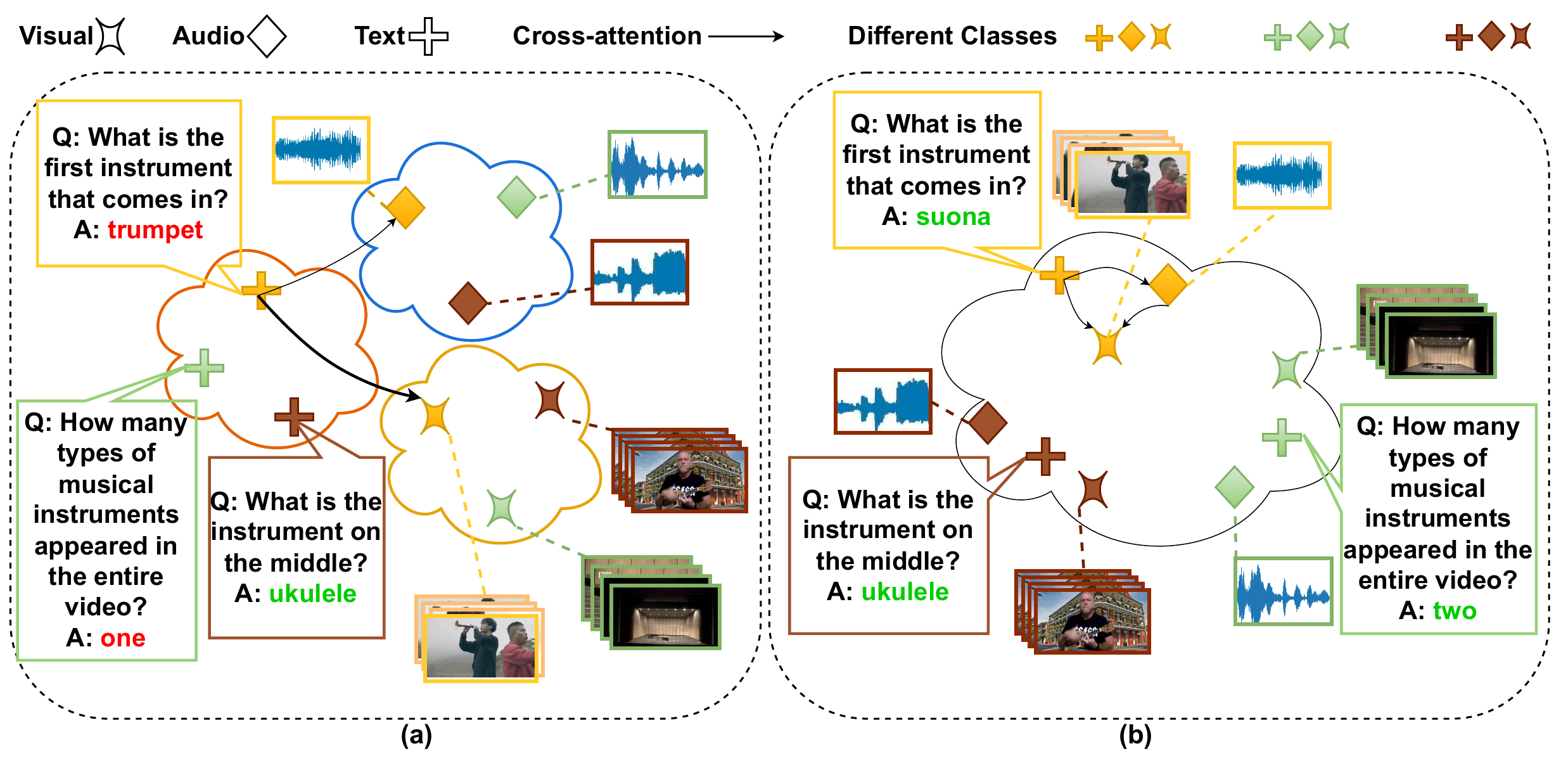}
   \vspace{-.6cm}
   \caption{Visualization of audio, text and visual representations learned by (a) state-of-the-art ST-AVQA \cite{li2022learning} and (b) CAD (ours).}  
   \label{fig:1}\vspace{-5mm}
\end{figure}

Audio-visual inputs have been used extensively in the literature to improve the performance of various tasks including video captioning \cite{iashin2020multi,tian2019audio,xu2017learning,nadeem2023sem}, speech recognition \cite{noda2015audio,hu2016temporal,afouras2018deep,song2022multimodal,shi2022robust}, speaker recognition \cite{sari2021multi,shon2019noise,qian2021audio,sell2018audio,gebru2017audio,chung2019said,ding2020self}, action recognition \cite{sun2022human,wang2016exploring,kazakos2019epic,xiao2020audiovisual,chen2022mm,gao2020listen,panda2021adamml,alfasly2022learnable,planamente2021cross,zhang2022audio}, emotion recognition \cite{cai2019multi,bedi2021multi,chauhan2020sentiment,delbrouck2020transformer,lv2021progressive,hazarika2018conversational}, sound localization \cite{owens2018audio,arandjelovic2018objects,senocak2018learning,hu2022mix,hu2021class}, saliency detection \cite{tavakoli2019dave,tsiami2020stavis,min2020multimodal,jain2021vinet,wang2021semantic}, event localization \cite{tian2018audio,lin2019dual,duan2021audio,lin2021exploring} and finally, question-answering \cite{alamri2019audio,hori2019joint,schwartz2019simple,geng2021dynamic,yun2021pano,li2022learning,shah2022audio}. However, training AV or multimodal methods is a challenging task and needs to be addressed on Spatial, Temporal, and Semantic levels \cite{wei2022learning}. For example, in Figure \ref{fig:1} - yellow class, the spatial location of the instrument can be determined using audio and visual inputs but to answer 'which comes first?', we also need to simultaneously hear the sound of the instrument and see it in visual input.

At this stage, any temporal misalignment between sound and visual appearance of the instrument can hamper learning. We also need to relate both the sound and appearance of the instrument with the label of the instrument on the Semantic level. In Figure \ref{fig:1}a, the state-of-the-art method ST-AVQA \cite{li2022learning} predicts the wrong answer which is 'trumpet' and our method in Figure \ref{fig:1}b predicts the right answer 'suona'.
Similarly, for green class, there are two instruments xylophone and piano and the question is about 'appeared' not 'played' as only the xylophone was played' and to answer correctly, the model requires Semantic level learning which ST-AVQA\cite{li2022learning} lacks.
ST-AVQA \cite{li2022learning} claims to assign larger weights to audio and visual segments which are more relevant to the asked question, by performing spatial grounding using audio and visual segments. This output is then temporally grounded with audio and question embedding. However, for scenarios as in the above examples, audio input is not available for both spatial and temporal groundings, which hampers the learning.
As shown in Figure \ref{fig:1}b, our contributions help align the same class modalities in a unified space with audio and text modalities used as queries as well as the chain of cross-attention, which helps in balancing AV information signals. In contrast, Figure \ref{fig:1}a represents the state-of-the-art method ST-AVQA\cite{li2022learning}, where different modalities are misaligned for the same class.

Other multimodal methods \cite{lu2019vilbert, tan2019lxmert, dosovitskiy2020image, sun2019videobert, sun2019learning, zhu2020actbert, akbari2021vatt} suffer from a major limitation of misaligned multi-modal information on Spatial and Temporal levels, leading to errors in the output. In AV learning, the audio and visual data are input separately to the network, and sampled at different rates. Due to the limitation of computing power and larger training time, the visual data is often more sparsely sampled as compared to audio data which requires less memory \cite{li2022learning}. This leads to two challenges: 1) misalignment between audio and visual streams; and 2) imbalance between audio and visual semantics within a context.
To achieve the best performance, ideally, the audio and visual information should be perfectly aligned and also, semantically balanced within a context. In this paper, we focus on the task of AVQA with multi-modal learning and address both the challenges in the existing methods.

Some methods \cite{lin2021exploring, akbari2021vatt, nagrani2021attention, morgado2020learning} use self-attention and cross-attention to look for the most co-related parts of the audio and visual features, however, the problem is only partially solved. Recently, transformer-based methods \cite{lu2019vilbert, tan2019lxmert, dosovitskiy2020image, sun2019videobert, sun2019learning, zhu2020actbert, akbari2021vatt} have become popular for their ability to use unlabeled data in a self-supervised manner by using contrastive learning which limits multi-modal class differentiation in the feature space. \cite{akbari2021vatt} employed the DropToken technique which drops features randomly to deal with the redundancy in audio and visual streams.  Other methods \cite{arandjelovic2017look, arandjelovic2018objects, korbar2018cooperative, morgado2021audio} synchronise audio and visual streams as part of the self-supervised learning and \cite{korbar2018cooperative} report a significant increase in performance on UCF101 and HMDB51 benchmarks. However, these methods do not resolve the problem of imbalance in AV semantics and misalignment between AV streams and do not work efficiently for more challenging datasets with dynamic scenarios \cite{li2022learning}.

In this paper, we aim to address the above challenges by introducing (1) a spatio-temporal Contextual block for visual features to mitigate the \textbf{AV spatial} misalignment; (2) \textbf{AV temporal} alignment of audio/visual features as a pre-training task in a self-supervised manner; and (3) a chain for the cross-attention modules in our architecture for creating \textbf{AV semantic} balance to handle dynamic scenarios.  

To summarise, we propose a \textbf{novel end-to-end network} for dynamic AVQA with three main contributions:
\begin{itemize}[topsep=0pt,partopsep=0pt,itemsep=0pt,parsep=0pt]
	\item Parameter-free contextual block to identify most context-related parts in the visual stream, reducing the spatial misalignment with the audio stream and reducing network complexity.
	\item AV fine temporal alignment uses self-supervised learning to pre-train the model for dynamically aligning audio and visual information over time. This helps the network understand and represent the temporal aspects of the data effectively.
	\item A chain of three cross-attention modules where initially audio and visual streams are cross-attended by question queries and then, the output of the audio cross-attention block queries the visual stream. This ensures that the network processes and balances the semantic (meaning-related) information from both modalities effectively.
\end{itemize}
Our contributions can be added to the existing methods to improve the performance without increasing any complexity (see Section \ref{ssec:ablation results}).

\section{Related Work}
\label{sec:relatedwork}
Previous research has primarily focused on question-answering tasks utilizing different modalities, such as audio, text, and vision, as input \cite{rajpurkar2016squad,weston2015towards,antol2015vqa,jang2017tgif,alamri2019audio,zhang2016yin}. In the context of visual question-answering (VQA) tasks \cite{goyal2017making,lu2016hierarchical}, the goal is to generate answers based solely on visual information. However, these methods do not incorporate spatio-temporal dynamics inherent in visual content. Recent advances address this limitation by enhancing the spatio-temporal reasoning ability through the integration of video-centered dynamics into the question-answering task \cite{yu2019activitynet,xiao2021next,li2019beyond,kim2020modality,fan2019heterogeneous}. This progress showcases the promising nature of AVQA as a developing research area \cite{li2022learning,yun2021pano}.

To facilitate AVQA research, multiple datasets have been introduced. The Pano-AVQA dataset \cite{yun2021pano} consists of 5.4k videos with questions based on dynamic audio, visual, and AV scenes. However, the AV questions in this dataset only cover existential and location categories. The MUSIC-AVQA dataset \cite{li2022learning} gives a comprehensive set of questions for AV scenes, including comparative, existential, counting, spatial, and temporal aspects in dynamic scenarios with Audio, visual, and AV modalities.

Li et al. \cite{li2022learning} employ spatial and temporal grounding techniques to perform the AVQA task using cross-attention. Interestingly, their ablations on the MUSIC-AVQA dataset demonstrate that using only audio as input yields better performance than using only visual in an AV scenario. This observation highlights a unique scenario where the same question (Q) is used as the query for both A and V, but V is not queried for A. As a result, the direct correlation between A and V appears to be missing, leading to the under-utilization of A-related information within the V modality. 

\noindent
\textbf{AV learning using Transformers} 
Transformer is a powerful architecture that has achieved exceptional performance on various language tasks including question-answering \cite{khan2022transformers}. Popular models based on transformers include BERT, RoBERTa, and GPT versions, with ChatGPT being particularly well-known \cite{devlin2018bert, liu2019roberta, radford2018improving, radford2019language, ouyang2022training}. Transformers typically follow a two-stage learning process. In the first stage, pre-training is performed using a large-scale dataset in either a supervised or self-supervised manner. This pre-training significantly enhances the performance of large transformer models in both vision and language tasks. Masking techniques have been introduced to process visual and textual streams during pre-training, yielding promising results \cite{devlin2018bert, he2022masked, li2020hero, liu2019roberta, xie2022simmim}.
Transformers employ a self-attention mechanism to improve their performance. Self-attention measures the relevance of different components within a sequence, such as patches of an image or words in a sentence, and models their interactions to optimize the output \cite{khan2022transformers}.

On the other hand, self-supervised learning (SSL) is used to train transformer models using unlabeled data, unlike supervised learning that relies on labeled data \cite{ericsson2022self}. SSL has the advantage of enabling training on large datasets that would be costly or time-consuming to label manually. This makes SSL a promising approach for developing AI systems that can tackle real-world problems. However, there are still challenges to overcome in SSL, including the development of more efficient algorithms and better methods for transferring learned representations across tasks \cite{ericsson2022self}.

Transformer models can be represented as graph neural networks, where self-attention processes the input as a fully-connected graph in a global fashion \cite{xu2022multimodal}. This enables transformers to effectively handle multiple modalities as nodes of the graph, leading to AV learning. AV learning involves tokenizing multiple modalities and representing them in a feature or embedding space. The design of this embedding space, which can have different granularity levels (e.g., fine or coarse), aims to establish class differentiation, often based on contrastive learning \cite{jin2022expectation}. However, contrastive learning faces challenges when dealing with AV tasks, as its performance relies on the quality and quantity of negative examples. Insufficient negative examples can prevent contrastive learning from converging and result in poor performance due to a lack of differentiation in the embedding space \cite{chen2020simple}.

To address the integration of multiple modalities in a single transformer architecture, cross-attention is used, allowing each modality to be processed with respect to the query of the other without significantly increasing computational cost. However, cross-attention tends to focus on global information, overlooking the fine-grained details within each modality, primarily due to the absence of a self-attention mechanism for individual modalities \cite{lin2020interbert}. Additionally, multimodal pre-training approaches utilize diverse large-scale multimodal datasets to train transformer-based models. These models, when trained on such datasets, outperform in multiple downstream tasks and demonstrate strong generalization abilities \cite{lu2019vilbert, tan2019lxmert, li2019visualbert, su2019vl, chen2020uniter}. The pre-training tasks aim to capture cross-modal interactions through either general or goal-oriented objectives. Unlike the alignment between co-occurring modalities in AV tasks, which has an inherent nature, transformer-based alignment techniques primarily leverage large amounts of data for vision-language tasks \cite{morgado2020learning, radford2021learning, jia2021scaling, xu2021videoclip, xu2022multimodal, lei2021less}. Preserving the alignment between modalities while mapping them into a shared embedding space remains a challenging task. Using contrastive learning to address this proves to be ineffective \cite{chen2020simple}.     


\section{Method}
This section gives an overview of the proposed method, followed by a problem statement and detailed explanations of three key components: (1) spatio-temporal visual Contextual block, (2) AV fine temporal alignment, and (3) Cross-Attention based network.

\begin{figure*}[t]
  \centering
   \includegraphics[width=\textwidth]{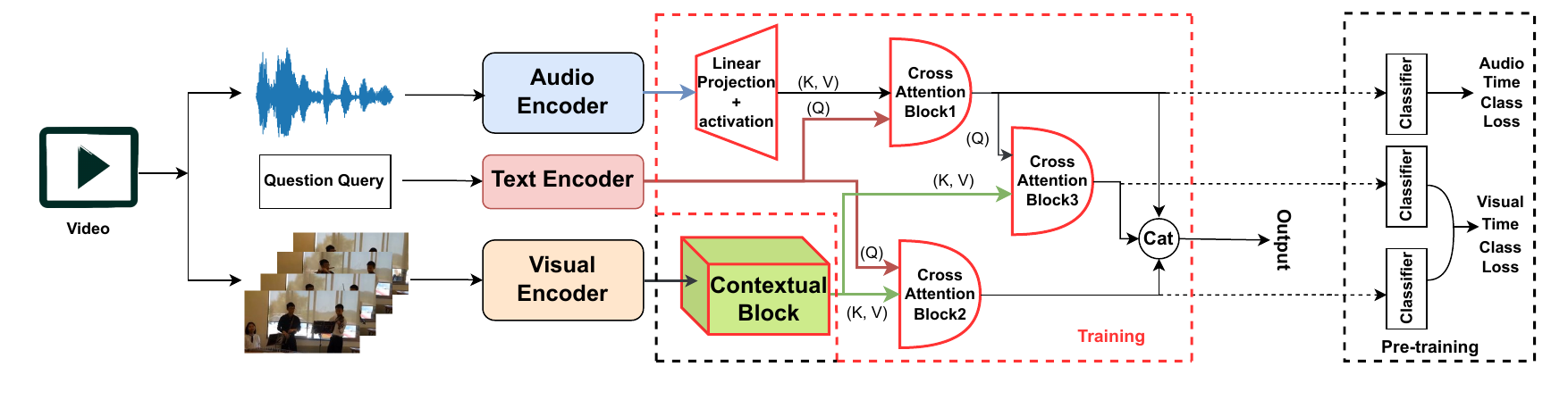}
   \vspace{-8mm}
   \caption{The overview of the proposed novel end-to-end \textbf{CAD} - \textbf{C}ontextual multi-modal \textbf{A}lignment for \textbf{D}ynamic AVQA. The novel parts of the network are highlighted in dotted lines.}  
   
   \label{fig:2}\vspace{-4.5mm}
\end{figure*}

\subsection{Overview:}
As shown in Figure \ref{fig:2}, our method takes audio and visual streams extracted from the video as input. The question query is encoded using \cite{pennington2014glove}.
Each input stream is passed through separate pre-trained backbone encoder networks -  one for audio \cite{kong2020panns} trained on AudioSet \cite{gemmeke2017audio} and another for visual input \cite{dosovitskiy2020image} trained on ImageNet \cite{deng2009imagenet}, to extract respective audio and visual features.
The visual features are fed into the parameter-free Contextual block to choose features relevant to the query. This decreases the size of the features reducing the overall complexity of the network.

The audio and visual (refined) features are input to a novel network setting of three cross-attention modules. These modules align the inputs and integrate information across modalities, and help in dealing with dynamic scenarios. The outputs of these modules are concatenated and passed through a classifier for output prediction for both training and testing. 
In the pre-training stage, each of these outputs is sent to separate classifiers, each responsible for predicting the time label.

\subsection{Problem Statement}
\label{ssec:prob_statement}
For challenging datasets such as MUSIC-AVQA \cite{li2022learning} with dynamic scenarios (types of questions including comparative, existential, counting, spatial and temporal aspects, in both audio, visual, and AV formats), existing networks suffer from two challenges.

The first one is that the audio and visual streams are often misaligned, which makes it difficult for the model to answer questions that require spatio-temporal reasoning.
Another challenge is that the semantic information in the audio and visual streams is imbalanced within a context.

To address these challenges, our method incorporates a pre-training task where the audio and visual inputs are segmented into 60 cues, each lasting for one second. These cues align the audio and visual streams temporally. The visual cues are input to the spatio-temporal stochastic Contextual block in pre-training as well as training, which aligns the audio and visual inputs spatially and also, reduces the overall complexity of the network, enhancing its efficiency. Our approach also employs three cross-attention modules as part of the pre-training process, which aligns and balances the audio and visual semantics.
Following the pre-training phase, we initialize our network with the pre-trained weights and proceed to train it in an end-to-end fashion, adopting a supervised learning approach. For a detailed description of the training procedure, please refer to Algorithm \textbf{1}.
\vspace{-3mm}

\begin{algorithm}
\label{algo:cad}
\caption{The proposed CAD method}\label{alg:cap}
\begin{algorithmic}
\Require $ $a$ \in A^{d_a}, $t$ \in T^{d_t}, $v$ \in V^{d_v} $ \Comment{Audio (a), Text (t) and Visual (v) features \\}
\emph{\bf{Initialize} model weights from the pre-trained weights\;}
\For{batch iteration n = 1, 2, ..., N}\\
\emph{\{Contextual block starts\}}
\If{training } \Comment{Sample 80\% of v features}
    \State $f_m \gets Mean($v$, dim=c)$ \Comment{c=channel}
    \State $M \gets Mask(f_m, th=Max(f_m)*0.9) $ 
    \State $C_f \gets sigmoid(f_m)$  \Comment{contextual features}
    \State $R \gets Random(C_f, M, select\_prob=0.9) $ 
    \State $ $v $ \gets $ v$*R $
\EndIf \\
\emph{\{Contextual block ends\}}
\State $a_t \gets CAB1(query=t, key=a, value=a)$
\State $v_t \gets CAB2(query=t, key=v, value=v)$
\State $v_{a_t} \gets CAB3(query=a_t, key=v, value=v)$
\emph{\{CAB = Cross Attention Block\}}
\State $answer \gets FC(Fuse\{a_t, v_t, v_{a_t}\}) $
\State $Loss \gets Cross\-Entropy(answer,ground\_truth)$

\EndFor

\end{algorithmic}

\end{algorithm}
\vspace{-3mm}

\subsection{Contextual Block}
\label{ssec:vd_block}
In this section, we explain our stochastic visual Contextual block, which plays a significant role in both the pre-training and training phases. In AV learning, the audio and visual streams contribute complementary information for the AVQA task. However, the learning process struggles in extracting spatial information from audio and visual streams concurrently, inadvertently limiting the network's ability to effectively learn from both streams \cite{li2022learning}.

The Contextual block extracts spatio-temporal visual context and allows the network to explicitly incorporate visual information most relevant for Spatial level learning. This facilitates better learning and improves the overall performance of the network when dealing with AV tasks. This block identifies contextually relevant regions in the visual input. To achieve this, we employ a series of steps that highlight these regions. At first, we randomly sample 80\% of the visual features for this process, as outlined in Algorithm \textbf{1}.

In the subsequent step, we average the visual features along the channel dimension. Then, we create a mask \(\mathnormal{M}\) with all the values greater than threshold \(\mathnormal{th}\) to be zero and otherwise, one. Similarly, we extract context \(\mathnormal{C}_{f}\) using sigmoid attention. Next, we introduce a stochastic selection between 90\% of either \(\mathnormal{M}\) or \(\mathnormal{C}_{f}\), for robustness, which is then masked out to zero. This helps to effectively reduce the complexity and training time. Further details are provided in the Appendix.
The feature space \(\mathcal{R}\)  is defined as:
\begin{equation}
\label{eqn:1}
\mathcal{R} = \{\{A^{d_a}, T^{d_t}, V^{d_v}\} |\, \forall
a \in A^{d_a}, t \in T^{d_t}, v \in V^{d_v}\} 
\end{equation}

Here, \( \mathnormal{A}^{d_a} \), \( \mathnormal{T}^{d_t} \) and \( \mathnormal{V}^{d_v} \) represent the audio, text, and visual feature spaces, respectively. Similary, \( \mathnormal{a} \), \( \mathnormal{t} \) and \( \mathnormal{v} \) denote the respective feature embeddings, with feature dimensions \( \mathnormal{d_a} \), \( \mathnormal{d_t} \) and \( \mathnormal{d_v} \).
The visual features are represented by 
\(\mathcal{V}^{[B * t * s]}\), where \(\mathnormal{B}\), \(\mathnormal{t}\) and \(\mathnormal{s}\) denote the batch size, number of time frames, and spatial size, respectively. The complexity of the network is directly influenced by \(\mathnormal{[t * s]}\).

The output of this block is input to two cross-attention modules sequentially which enables the network to focus on the most relevant information for Spatial level learning.

\subsection{AV Fine Temporal Alignment as pre-training}
\label{ssec:mmfa}
\begin{figure*}[t]
  \centering
   \includegraphics[width=\textwidth]{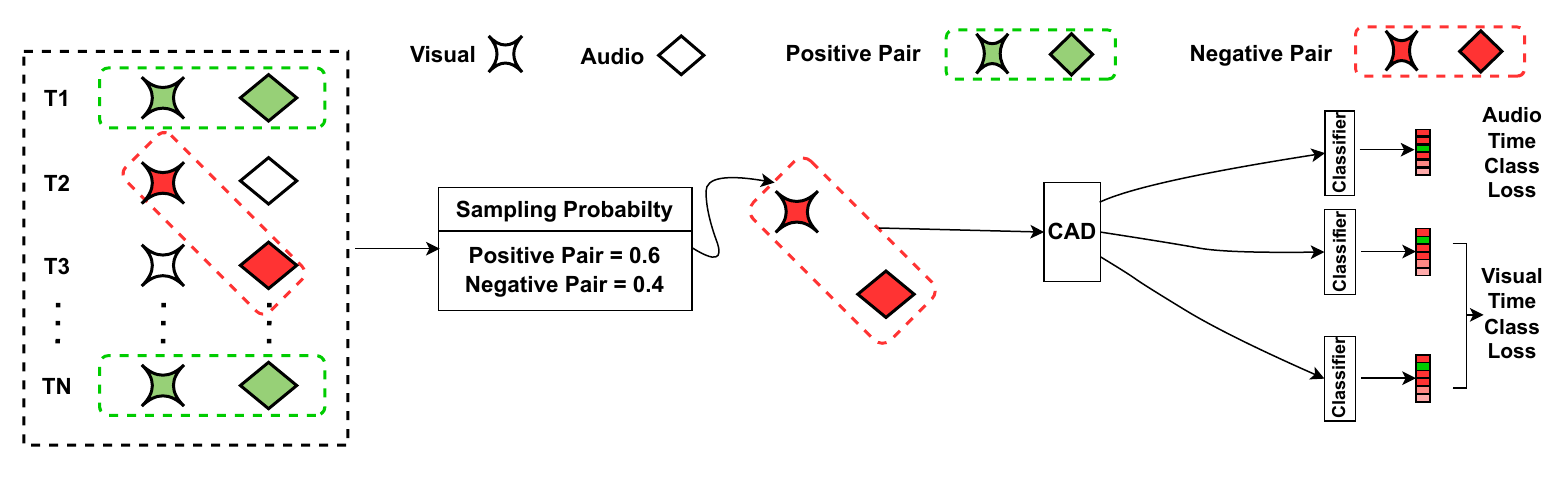}
   \vspace{-10mm}
   \caption{The overview of the proposed AV Fine Temporal Alignment where encoded audio-visual positive and negative pairs are sampled with a probability of 0.6 and 0.4 respectively before passing these through our CAD network. Time labels (N=60) for audio and visual are predicted separately via Softmax classifiers.}  
   \label{fig:4}\vspace{-5mm}
\end{figure*}

The contextual block takes as input visual features, and it outputs a new set of features that are more spatially relevant to the audio stream. This makes it easier for the model to learn the interactions between the audio and visual streams. 
In this section, we pre-train our neural network to classify the time label or class of both audio and visual streams for better Temporal level learning.
We propose an objective for the pre-training task to temporally align the audio and visual streams for question answering. 
We input audio and visual features to the network as a combination of positive and negative pairs. Positive pairs are pairs of audio and visual features that are aligned, and negative pairs are pairs of audio and visual features that are not aligned as shown in Figure \ref{fig:4}. \\
This helps the network to learn to distinguish between aligned and non-aligned pairs. Audio and visual features are input stochastically to the network in a combination of positive and negative pairs with an overall share of 60\% and 40\% respectively which is selected after thorough experimentation provided in the Appendix.
As shown in Figures \ref{fig:2} and \ref{fig:4}, three classifiers are added at the end of the network to predict audio and visual time labels. Visual stream is predicted by two classifiers with both text and attended (output of audio cross-attention block) audio as queries. We use the cross-entropy loss for all three classifiers. The only difference in the training phase is removing the three classifiers and the concatenation of outputs of the cross-attention modules prior to it  (see Figure \ref{fig:2}). 

\noindent

\subsection{Cross Attention Blocks}
Our approach aims to effectively address the semantic information imbalance in the audio and visual streams, by incorporating three cross-attention blocks within the network, as seen in Figure \ref{fig:2}. 
Two cross-attention blocks take the audio features and post-Contextual block visual features as input keys and values while utilizing question features as queries. These blocks enable the network to capture relevant information from both the audio and visual modalities and align it with the query, facilitating a comprehensive understanding of the question.
The third cross-attention block incorporates visual features as keys and values, with queries sourced from the output of the audio cross-attention module. This configuration ensures that information from the audio stream is effectively propagated to the visual stream, allowing for enhanced integration and alignment of features. By incorporating all three cross-attention blocks, we provide a robust framework that covers all the aforementioned dynamic scenarios, resulting in improved overall performance. Algorithm \textbf{2} gives the technical details about the cross-attention block. 
\vspace{-3mm}
\begin{algorithm}
\label{algo:cab}
\caption{The proposed Cross-Attention Blocks}\label{alg:cap}
\begin{algorithmic}
\Require $ $k$, $v$, $q$ $ \Comment{key (k), value(v) and query (q) \\}
\emph{\bf{Initialize} weights from the pre-trained weights\;}
    \State $f_a \gets Multi\_Headed\_Attention($q$, $k$, $v$)$ 
    \State $fc1 \gets FC(f_a) $ \Comment{fully connected layer (512,512)}
    \State $r1 \gets relu(fc1) $ \Comment{Activation}
    \State $fc2 \gets FC(r1) $ \Comment{fully connected layer (512,512)}
    \State $r2 \gets relu(fc2) $ \Comment{Activation}
    \State $f_f \gets f_a + r2  $ \Comment{Addition}
    \State $f_n \gets norm(f_f)  $ \Comment{Layer Norm\\}
\emph{\bf{return} $f_n$ \;}

\end{algorithmic}

\end{algorithm}
\vspace{-4mm}
\subsection{End-to-end CAD}
In this section, we explain the end-to-end architecture employed in our method.
The outputs from three cross-attention blocks are concatenated together and fed into a fully-connected layer that learns the answer embedding. This enables the network to make predictions on the most likely answer for a given input.
\begin{equation}
\label{eqn:2}
\mathcal{L}_{avqa} = -\displaystyle\sum_{n=1} ^{N} \mathnormal{A}_{n} \mathnormal{log} \mathnormal{P}_{n}
\end{equation}
\(\mathnormal{A}_{n}\) is the actual answer embedding and \(\mathnormal{P}_{n}\) is the output of the classifier and \(\mathcal{L}_{avqa}\) is the AVQA loss. 
During the training phase, we employ the cross-entropy loss function to train our network. This loss function effectively measures the dissimilarity between the predicted and the ground truth labels, facilitating the optimization process. It is worth noting that the network, prior to the fusion, is initialized with weights obtained from the pre-training phase. This initialization provides a starting point that already incorporates valuable knowledge obtained during pre-training, allowing for more efficient learning during the training phase.

\section{Experimental Results and Implementation}

\subsection{Dataset} We train and test our model on MUSIC-AVQA \cite{li2022learning}, a large-scale dataset that contains question-answer pairs for audio, visual, and AV questions about musical performance. This dataset contains 9,290 YouTube videos, 45,867 question-answer pairs and 9 types of diverse, complex and dynamic AV questions. We employ the ACAV100M \cite{lee2021acav100m} dataset for the pre-training task. We detail our pre-training stage dataset sampling, preprocessing, and labeling strategy in the Appendix. 


\begin{table*}[bp]
\centering
\resizebox{\textwidth}{!}{%
\begin{tabular}{*{1}c|*{1}c|*{13}c}
\hline
Task & Method-Year & \multicolumn{3}{c|}{Audio}                                                                                                    & \multicolumn{3}{c|}{Visual} & \multicolumn{6}{c|}{Audio-Visual} & Avg                                                                                                  \\ \cline{3-14} 
                        &                                       &\multicolumn{1}{c}{Count}    & \multicolumn{1}{c}{Comp}    & \multicolumn{1}{c|}{Avg}    & \multicolumn{1}{c}{Count} & \multicolumn{1}{c}{Localis}                          & \multicolumn{1}{c|}{Avg}    & \multicolumn{1}{c}{Exist}    & \multicolumn{1}{c}{Localis}    & \multicolumn{1}{c}{Count} & \multicolumn{1}{c}{Comp} & \multicolumn{1}{c}{Temp}    & \multicolumn{1}{c|}{Avg} &                    \\ \hline 
AudioQA                 & \multicolumn{1}{c|}{FCNLSTM \cite{fayek2020temporal}-TASLP2019}          & \multicolumn{1}{c}{70.45}          & \multicolumn{1}{c}{66.22} & \multicolumn{1}{c|}{68.88}            & \multicolumn{1}{c}{63.89}          & \multicolumn{1}{c}{46.74}          & \multicolumn{1}{c|}{55.21}          & \multicolumn{1}{c}{82.01}         & \multicolumn{1}{c}{46.28} & \multicolumn{1}{c}{59.34}& \multicolumn{1}{c}{62.15}& \multicolumn{1}{c}{47.33} & \multicolumn{1}{c|}{60.06} & 60.34           \\  

                 & \multicolumn{1}{c|}{CONVLSTM \cite{fayek2020temporal}-TASLP2019}             & \multicolumn{1}{c}{74.07}          & \multicolumn{1}{c}{68.89}          & \multicolumn{1}{c|}{72.15} & \multicolumn{1}{c}{67.47}            & \multicolumn{1}{c}{54.56}          & \multicolumn{1}{c|}{60.94}          & \multicolumn{1}{c}{82.91}          & \multicolumn{1}{c}{50.81}         & \multicolumn{1}{c}{63.03} & \multicolumn{1}{c}{60.27}& \multicolumn{1}{c}{51.58}& \multicolumn{1}{c|}{62.24} & 63.65
\\ \hline 
VisualQA                 & \multicolumn{1}{c|}{GRU \cite{antol2015vqa}-ICCV2015}             & \multicolumn{1}{c}{72.21}          & \multicolumn{1}{c}{66.89}          & \multicolumn{1}{c|}{70.24} & \multicolumn{1}{c}{67.72}            & \multicolumn{1}{c}{70.11}          & \multicolumn{1}{c|}{68.93}          & \multicolumn{1}{c}{81.71}          & \multicolumn{1}{c}{59.44}         & \multicolumn{1}{c}{62.64} & \multicolumn{1}{c}{61.88}& \multicolumn{1}{c}{60.07}& \multicolumn{1}{c|}{65.18} & 67.07           \\  

                 & \multicolumn{1}{c|}{BiLSTM Attn \cite{zhou2016attention}-ACL2016}             & \multicolumn{1}{c}{70.35}          & \multicolumn{1}{c}{47.92}          & \multicolumn{1}{c|}{62.05} & \multicolumn{1}{c}{64.64}            & \multicolumn{1}{c}{64.33}          & \multicolumn{1}{c|}{64.48}          & \multicolumn{1}{c}{78.39}          & \multicolumn{1}{c}{45.85}         & \multicolumn{1}{c}{56.91} & \multicolumn{1}{c}{53.09}& \multicolumn{1}{c}{49.76}& \multicolumn{1}{c|}{57.10} & 59.92           \\ 

                 & \multicolumn{1}{c|}{HCAttn \cite{lu2016hierarchical}-NeurIPS2016}             & \multicolumn{1}{c}{70.25}          & \multicolumn{1}{c}{54.91}          & \multicolumn{1}{c|}{64.57} & \multicolumn{1}{c}{64.05}            & \multicolumn{1}{c}{66.37}          & \multicolumn{1}{c|}{65.22}          & \multicolumn{1}{c}{79.10}          & \multicolumn{1}{c}{49.51}         & \multicolumn{1}{c}{59.97} & \multicolumn{1}{c}{55.25}& \multicolumn{1}{c}{56.43}& \multicolumn{1}{c|}{60.19} & 62.30           \\  

                 & \multicolumn{1}{c|}{MCAN \cite{yu2019deep}-CVPR2019}             & \multicolumn{1}{c}{77.50}          & \multicolumn{1}{c}{55.24}          & \multicolumn{1}{c|}{69.25} & \multicolumn{1}{c}{71.56}            & \multicolumn{1}{c}{70.93}          & \multicolumn{1}{c|}{71.24}          & \multicolumn{1}{c}{80.40}          & \multicolumn{1}{c}{54.48}         & \multicolumn{1}{c}{64.91} & \multicolumn{1}{c}{57.22}& \multicolumn{1}{c}{47.57}& \multicolumn{1}{c|}{61.58} & 65.49
\\ \hline
VideoQA                 & \multicolumn{1}{c|}{PSAC \cite{li2019beyond}-AAAI2019}             & \multicolumn{1}{c}{75.64}          & \multicolumn{1}{c}{66.06}          & \multicolumn{1}{c|}{72.09} & \multicolumn{1}{c}{68.64}            & \multicolumn{1}{c}{69.79}          & \multicolumn{1}{c|}{69.22}          & \multicolumn{1}{c}{77.59}          & \multicolumn{1}{c}{55.02}         & \multicolumn{1}{c}{63.42} & \multicolumn{1}{c}{61.17}& \multicolumn{1}{c}{59.47}& \multicolumn{1}{c|}{63.52} & 66.54           \\ 

                 & \multicolumn{1}{c|}{HME \cite{fan2019heterogeneous}-CVPR2019}             & \multicolumn{1}{c}{74.76}          & \multicolumn{1}{c}{63.56}          & \multicolumn{1}{c|}{70.61} & \multicolumn{1}{c}{67.97}            & \multicolumn{1}{c}{69.46}          & \multicolumn{1}{c|}{68.76}          & \multicolumn{1}{c}{80.30}          & \multicolumn{1}{c}{53.18}         & \multicolumn{1}{c}{63.19} & \multicolumn{1}{c}{62.69}& \multicolumn{1}{c}{59.83}& \multicolumn{1}{c|}{64.05} & 66.45           \\  

                 & \multicolumn{1}{c|}{HCRN \cite{le2020hierarchical}-CVPR2020}             & \multicolumn{1}{c}{68.59}          & \multicolumn{1}{c}{50.92}          & \multicolumn{1}{c|}{62.05} & \multicolumn{1}{c}{64.39}            & \multicolumn{1}{c}{61.81}          & \multicolumn{1}{c|}{63.08}          & \multicolumn{1}{c}{54.47}          & \multicolumn{1}{c}{41.53}         & \multicolumn{1}{c}{53.38} & \multicolumn{1}{c}{52.11}& \multicolumn{1}{c}{47.69}& \multicolumn{1}{c|}{50.26} & 55.73
                 \\ \hline
AVQA                 & \multicolumn{1}{c|}{AVSD \cite{schwartz2019simple}-CVPR2019}             & \multicolumn{1}{c}{72.41}          & \multicolumn{1}{c}{61.90}          & \multicolumn{1}{c|}{68.52} & \multicolumn{1}{c}{67.39}            & \multicolumn{1}{c}{74.19}          & \multicolumn{1}{c|}{70.83}          & \multicolumn{1}{c}{81.61}          & \multicolumn{1}{c}{58.79}         & \multicolumn{1}{c}{63.89} & \multicolumn{1}{c}{61.52}& \multicolumn{1}{c}{61.41}& \multicolumn{1}{c|}{65.49} & 67.44           \\  

                 & \multicolumn{1}{c|}{PanoAVQA \cite{yun2021pano}-ICCV2021}             & \multicolumn{1}{c}{74.36}          & \multicolumn{1}{c}{64.56}          & \multicolumn{1}{c|}{70.73} & \multicolumn{1}{c}{69.39}            & \multicolumn{1}{c}{75.65}          & \multicolumn{1}{c|}{72.56}          & \multicolumn{1}{c}{81.21}          & \multicolumn{1}{c}{59.33}         & \multicolumn{1}{c}{64.91} & \multicolumn{1}{c}{64.22}& \multicolumn{1}{c}{63.23}& \multicolumn{1}{c|}{66.64} & 68.93
                     \\ 

                 & \multicolumn{1}{c|}{ST-AVQA \cite{li2022learning}-CVPR2022}             & \multicolumn{1}{c}{78.18}          & \multicolumn{1}{c}{67.05}          & \multicolumn{1}{c|}{74.06} & \multicolumn{1}{c}{71.56}            & \multicolumn{1}{c}{76.38}          & \multicolumn{1}{c|}{74.00}          & \multicolumn{1}{c}{81.81}          & \multicolumn{1}{c}{64.51}         & \multicolumn{1}{c}{70.80} & \multicolumn{1}{c}{66.01}& \multicolumn{1}{c}{63.23}& \multicolumn{1}{c|}{69.54} & 71.52 \\  \cline{2-15 \vspace{0.5mm}} 
                 & \multicolumn{1}{c|}{\textbf{CAD (Ours)}}             & \multicolumn{1}{c}{\textbf{82.91}}          & \multicolumn{1}{c}{\textbf{73.34}}          & \multicolumn{1}{c|}{\textbf{78.13}} & \multicolumn{1}{c}{\textbf{78.21}}            & \multicolumn{1}{c}{\textbf{81.19}}          & \multicolumn{1}{c|}{\textbf{79.70}}          & \multicolumn{1}{c}{\textbf{83.42}}          & \multicolumn{1}{c}{\textbf{73.97}}         & \multicolumn{1}{c}{\textbf{76.37}} & \multicolumn{1}{c}{\textbf{74.88}}& \multicolumn{1}{c}{\textbf{76.16}}& \multicolumn{1}{c|}{\textbf{76.96}} & \textbf{78.26} 
\\ \hline

\end{tabular}%
}
\parbox[t]{\textwidth}{\vspace{-0.1mm}\caption{Comparison against state-of-the-art methods. The best results in each category are in bold.} 
\label{table:1}}\vspace{-8mm}
\end{table*}

\subsection{Implementation} The audio input is sampled at 32kHz, which is a standard sampling rate for audio. We use PANNs \cite{kong2020panns} to extract features from audio data.
The visual input is sampled at 15 frames per second in the pre-training and training. We use ViT \cite{dosovitskiy2020image}, which is a transformer-based network, to extract features from images. 
The textual/question input is embedded using GLoVE \cite{pennington2014glove}. The hidden dimension and cross-attention dimension size are both set to 512.
The cross-attention modules employ 8 heads. The other training parameters are a learning rate of 0.0001, 25 epochs, a batch size of 64, and the ADAM optimizer. The training is done using one NVIDIA GeForce RTX 2080 Ti GPU.  We use the same parameters for pre-training except the epochs are 10. Further details are provided in the Appendix.  

\subsection{Results and Comparison}

\noindent
\textbf{Quantitative Evaluation.} 
We compare the quantitative performance of the proposed method with the state-of-the-art approaches. For MUSIC-AVQA \cite{li2022learning} dataset, the evaluation encompasses three multi-modal scenes, namely audio, visual, and audio-visual, and covers a total of nine different types of questions (refer to Table \ref{table:1}).
 For a fair comparison, we use the prediction accuracy of the answer as the evaluation metric where the answer vocabulary contains 42 possible answers. Notably, our method showcases superior performance across all categories and question types when compared to state-of-the-art techniques. 
 Specifically, in the AV scene, our method achieves an improvement of 10.7\%, followed by 7.7\% and 5.5\% improvements in the V and A scenes, respectively. These results affirm the effectiveness of our approach in tackling diverse audio-visual scenarios and addressing various types of questions.
 Our method showcases improvements, particularly in addressing AV temporal, localization, and comparative questions, where we observe enhancements of 20.5\%, 14.7\%, and 13.4\% respectively. This demonstrates the effectiveness of our approach in capturing and understanding temporal dynamics, accurately localizing elements, and facilitating comparative reasoning within AV contexts. Additionally, we achieve notable improvements of 9.3\% and 9.4\% in V counting and A comparative questions, further highlighting the versatility and robustness of our method across different modalities. 
 While the improvement is relatively lower in AV existential questions, our method still achieves a modest enhancement of 2\%, indicating its capability to handle and reason about the existence of AV elements. It is worth mentioning that our method belongs to the AVQA task in Table \ref{table:1} and all the listed methods are trained on the AVQA benchmark dataset as reported in \cite{li2022learning}. These tasks are categorized based on the input modality or modalities. We also train and test our CAD method on MSRVTT-QA\cite{xu2017video} and ActvityNet-QA\cite{yu2019activitynet} benchmark datasets and demonstrate significant improvement in comparison to the existing methods by optimizing AV learning as shown in Table \ref{table:3}. These results provide compelling evidence of the efficacy and applicability of our method in various question types, with substantial advancements in addressing specific challenges related to temporal, spatial, and comparative aspects in the AV domain. 

\begin{figure*}[bp]
  \centering
   \includegraphics[width=\textwidth]{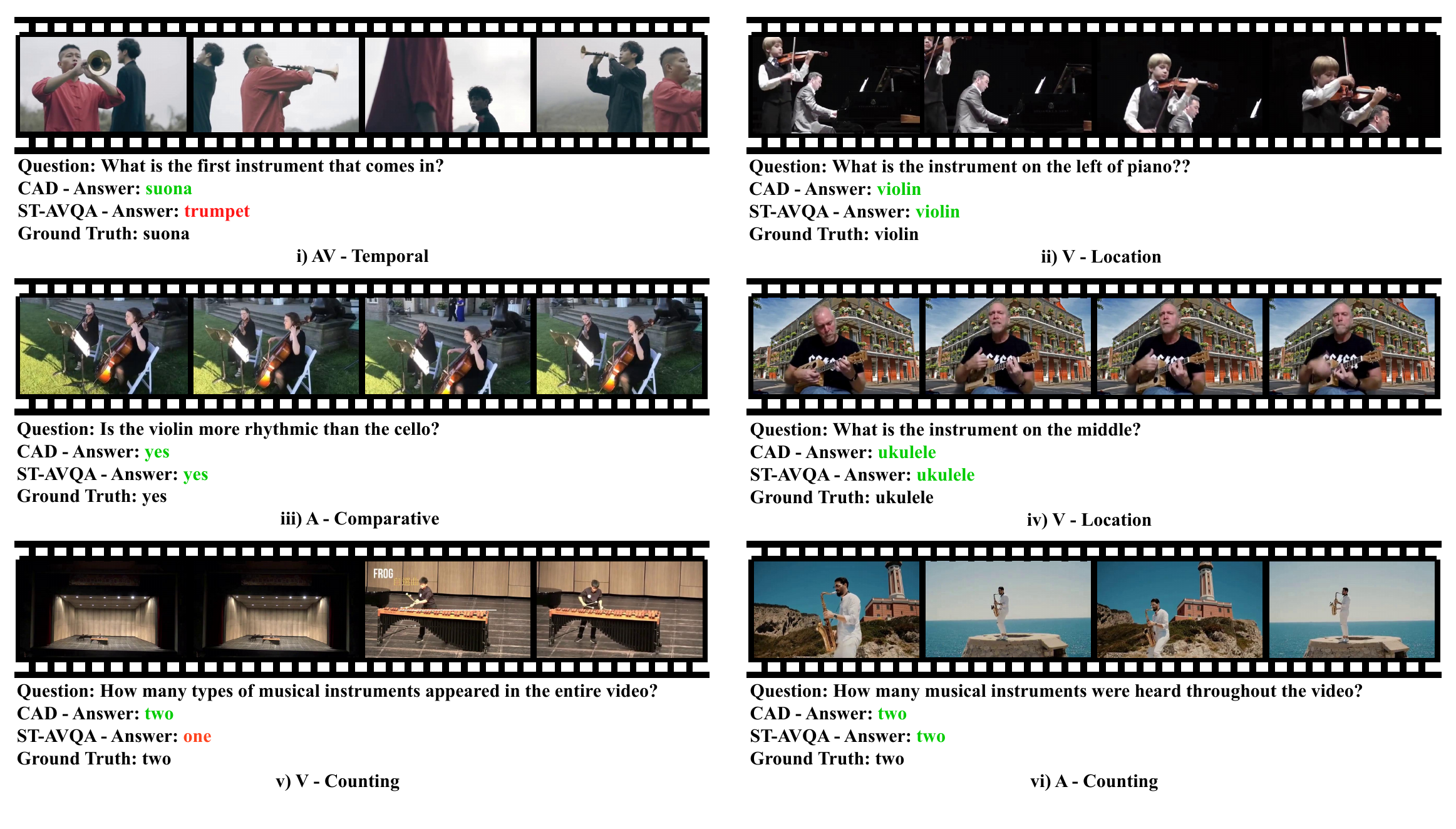}
   \caption{Qualitative results comparison against ground-truth and ST-AVQA \cite{li2022learning}.} 
   \label{fig:3} 
\end{figure*}

\noindent
\textbf{Qualitative Evaluation.} In Figure \ref{fig:3}, we demonstrate the results of our method and do a comparison with the state-of-the-art ST-AVQA \cite{li2022learning} and the ground-truth. In terms of audio-visual (AV) and visual (V) categories, our method demonstrates better results than the ST-AVQA \cite{li2022learning}(see subfigures i, v in Figure \ref{fig:3}). 
 
\vspace{-3mm}

\begin{table}[H]
\centering
\resizebox{\linewidth}{!}{%
\begin{tabular}{*{1}c|*{1}c|*{1}c|*{1}c}
\hline
Modality & Method & MSRVTT-QA & ActivityNet-QA                                                                                                  \\ \hline 

V+Q            & \multicolumn{1}{c|}{QueST \cite{jiang2020divide}}             & 34.6 & -
                     \\ 
\cline{2-4 \vspace{0.5mm}} 
& \multicolumn{1}{c|}{ClipBERT \cite{lei2021less}}             & 37.4 & -
                     \\ 
\cline{2-4 \vspace{0.5mm}} 
& \multicolumn{1}{c|}{JustAsk \cite{yang2021just}}             & 41.5 & 38.9
                     \\ 
\cline{2-4 \vspace{0.5mm}}
& \multicolumn{1}{c|}{MV-GPT \cite{seo2022end}}             & 41.7 & 39.1
                     \\ 
\cline{2-4 \vspace{0.5mm}}
& \multicolumn{1}{c|}{MERLOT \cite{zellers2021merlot}}             & 43.1 & 41.4
                     \\ 
\cline{2-4 \vspace{0.5mm}}
& \multicolumn{1}{c|}{SINGULARITY \cite{lei2022revealing}}             & 43.5 & 43.1
                     \\ 
\cline{2-4 \vspace{0.5mm}}
& \multicolumn{1}{c|}{VIOLET \cite{fu2023empirical}}             & 44.5 & -
                     \\ 
\cline{2-4 \vspace{0.5mm}}
& \multicolumn{1}{c|}{FrozenBiLM \cite{yang2022zero}}             & 47.0 & 43.2
                     \\ 
\cline{2-4 \vspace{0.5mm}} 
& \multicolumn{1}{c|}{Flamingo \cite{alayrac2022flamingo}}             & 47.4 & -
                     \\ 
\cline{2-4 \vspace{0.5mm}}
& \multicolumn{1}{c|}{CAD (Ours)}             & \underline{47.53} & \underline{46.92}
                     \\                      
\cline{1-4 \vspace{0.5mm}}
    V+A+Q             & \multicolumn{1}{c|}{\textbf{CAD (Ours)}}             & \textbf{49.06} & \textbf{48.81}
\\ \hline

\end{tabular}%
}
\parbox[t]{\linewidth}{\vspace{-0.2mm}\caption{Comparison against state-of-the-art methods on MSRVTT-QA\cite{xu2017video} and ActivityNet-QA\cite{yu2019activitynet} datasets. The best results in each category are in bold.} 
\label{table:3}\vspace{-8mm}}
\end{table}

\begin{table*}[t!]
\centering
\resizebox{\textwidth}{!}{%
\begin{tabular}{*{1}c|*{13}c}
\hline
Task & \multicolumn{3}{c|}{Audio}  & \multicolumn{3}{c|}{Visual} & \multicolumn{6}{c|}{Audio-Visual} & Avg                                                                                                  \\ \cline{2-13} 
                        &\multicolumn{1}{c}{Count}    & \multicolumn{1}{c}{Comp}    & \multicolumn{1}{c|}{Avg}    & \multicolumn{1}{c}{Count} & \multicolumn{1}{c}{Localis}                          & \multicolumn{1}{c|}{Avg}    & \multicolumn{1}{c}{Exist}    & \multicolumn{1}{c}{Localis}    & \multicolumn{1}{c}{Count} & \multicolumn{1}{c}{Comp} & \multicolumn{1}{c}{Temp}    & \multicolumn{1}{c|}{Avg} &                    \\ \hline 
w/o Pre-training                           & \multicolumn{1}{c}{79.17}          & \multicolumn{1}{c}{69.81} & \multicolumn{1}{c|}{74.49}            & \multicolumn{1}{c}{74.53}          & \multicolumn{1}{c}{78.06}          & \multicolumn{1}{c|}{76.30}          & \multicolumn{1}{c}{81.91}         & \multicolumn{1}{c}{67.47} & \multicolumn{1}{c}{72.94}& \multicolumn{1}{c}{70.23}& \multicolumn{1}{c}{67.88} & \multicolumn{1}{c|}{72.09} & 74.29           \\ 

w/o Contextual block               &  \multicolumn{1}{c}{81.13}          & \multicolumn{1}{c}{70.05}          & \multicolumn{1}{c|}{75.59} & \multicolumn{1}{c}{75.91}            & \multicolumn{1}{c}{79.62}          & \multicolumn{1}{c|}{77.77}          & \multicolumn{1}{c}{82.24}          & \multicolumn{1}{c}{70.41}         & \multicolumn{1}{c}{73.14} & \multicolumn{1}{c}{71.37}& \multicolumn{1}{c}{71.26}& \multicolumn{1}{c|}{73.68} & 75.68           \\ 

                 w 3CA only              & \multicolumn{1}{c}{78.8}          & \multicolumn{1}{c}{69.07}          & \multicolumn{1}{c|}{73.94} & \multicolumn{1}{c}{74.11}            & \multicolumn{1}{c}{76.98}          & \multicolumn{1}{c|}{75.55}          & \multicolumn{1}{c}{81.05}          & \multicolumn{1}{c}{66.53}         & \multicolumn{1}{c}{71.1} & \multicolumn{1}{c}{69.28}& \multicolumn{1}{c}{66.41}& \multicolumn{1}{c|}{70.87} & 73.45 \\

                 w 2CA            & \multicolumn{1}{c}{82.04}          & \multicolumn{1}{c}{73.21}          & \multicolumn{1}{c|}{77.63} & \multicolumn{1}{c}{73.34}            & \multicolumn{1}{c}{76.68}          & \multicolumn{1}{c|}{75.01}          & \multicolumn{1}{c}{82.21}          & \multicolumn{1}{c}{66.9}         & \multicolumn{1}{c}{72.13} & \multicolumn{1}{c}{69.74}& \multicolumn{1}{c}{64.99}& \multicolumn{1}{c|}{71.19} & 74.61          \\ 

                  w 4CA            & \multicolumn{1}{c}{80.04}          & \multicolumn{1}{c}{71.57}          & \multicolumn{1}{c|}{75.81} & \multicolumn{1}{c}{76.01}            & \multicolumn{1}{c}{80.11}          & \multicolumn{1}{c|}{78.06}          & \multicolumn{1}{c}{81.35}          & \multicolumn{1}{c}{70.25}         & \multicolumn{1}{c}{73.13} & \multicolumn{1}{c}{71.7}& \multicolumn{1}{c}{72.04}& \multicolumn{1}{c|}{73.69} & 75.85          \\ 
									
                w ST-AVQA \cite{li2022learning} + Pre-training              & \multicolumn{1}{c}{79.04}          & \multicolumn{1}{c}{68.13}          & \multicolumn{1}{c|}{73.59} & \multicolumn{1}{c}{72.33}            & \multicolumn{1}{c}{76.51}          & \multicolumn{1}{c|}{74.42}          & \multicolumn{1}{c}{81.99}          & \multicolumn{1}{c}{65.41}         & \multicolumn{1}{c}{71.45} & \multicolumn{1}{c}{68.02}& \multicolumn{1}{c}{64.61}& \multicolumn{1}{c|}{70.30} & 72.77 \\
                												
                w ST-AVQA \cite{li2022learning} + Contextual block              & \multicolumn{1}{c}{78.56}          & \multicolumn{1}{c}{67.92}          & \multicolumn{1}{c|}{73.24} & \multicolumn{1}{c}{71.95}            & \multicolumn{1}{c}{76.62}          & \multicolumn{1}{c|}{74.29}          & \multicolumn{1}{c}{81.9}          & \multicolumn{1}{c}{65.13}         & \multicolumn{1}{c}{71.31} & \multicolumn{1}{c}{67.55}& \multicolumn{1}{c}{64.89}& \multicolumn{1}{c|}{70.16} & 72.56 \\  
                
                w Q              & \multicolumn{1}{c}{71.11}          & \multicolumn{1}{c}{63.49}          & \multicolumn{1}{c|}{67.30} & \multicolumn{1}{c}{54.06}            & \multicolumn{1}{c}{56.44}          & \multicolumn{1}{c|}{55.25}          & \multicolumn{1}{c}{70.32}          & \multicolumn{1}{c}{56.64}         & \multicolumn{1}{c}{61.33} & \multicolumn{1}{c}{58.77}& \multicolumn{1}{c}{50.17}& \multicolumn{1}{c|}{59.45} & 60.67 \\

                w A + Q              & \multicolumn{1}{c}{81.51}          & \multicolumn{1}{c}{72.18}          & \multicolumn{1}{c|}{76.85} & \multicolumn{1}{c}{61.42}            & \multicolumn{1}{c}{60.2}          & \multicolumn{1}{c|}{60.81}          & \multicolumn{1}{c}{78.89}          & \multicolumn{1}{c}{61.15}         & \multicolumn{1}{c}{68.3} & \multicolumn{1}{c}{67.56}& \multicolumn{1}{c}{59.53}& \multicolumn{1}{c|}{67.09} & 68.25 \\
												
                w V + Q              & \multicolumn{1}{c}{74.57}          & \multicolumn{1}{c}{67.24}          & \multicolumn{1}{c|}{70.91} & \multicolumn{1}{c}{74.66}            & \multicolumn{1}{c}{76.37}          & \multicolumn{1}{c|}{75.52}          & \multicolumn{1}{c}{80.12}          & \multicolumn{1}{c}{69.4}         & \multicolumn{1}{c}{64.53} & \multicolumn{1}{c}{63.68}& \multicolumn{1}{c}{62.17}& \multicolumn{1}{c|}{67.98} & 71.47 \\
                 \hline 
                 \textbf{w A + V + Q - (CAD - Ours)}            & \multicolumn{1}{c}{\textbf{82.91}}          & \multicolumn{1}{c}{\textbf{73.34}}          & \multicolumn{1}{c|}{\textbf{78.13}} & \multicolumn{1}{c}{\textbf{78.21}}            & \multicolumn{1}{c}{\textbf{81.19}}          & \multicolumn{1}{c|}{\textbf{79.70}}          & \multicolumn{1}{c}{\textbf{83.42}}          & \multicolumn{1}{c}{\textbf{73.97}}         & \multicolumn{1}{c}{\textbf{76.37}} & \multicolumn{1}{c}{\textbf{74.88}}& \multicolumn{1}{c}{\textbf{76.16}}& \multicolumn{1}{c|}{\textbf{76.96}} & \textbf{78.26} 
\\ \hline
\end{tabular}%
}\vspace{-0mm}
\parbox[t]{\textwidth}{\caption{Ablation results. The best results in each category are in bold.} 
\label{table:2}\vspace{-17mm}}
\end{table*}

\noindent
Both of these results demonstrate the ability of our method, as shown in quantitative evaluation, to perform better than the state-of-the-art in AV and V scenes and also, perform equally well in audio (A) category.\\ In Figure \ref{fig:3}\textbf{i}, the state-of-the-art method ST-AVQA \cite{li2022learning} predicts the wrong answer which is 'trumpet' and our method predicts the right answer 'suona'. The spatial location of the 'suona' can be determined using audio and visual inputs but to answer which comes first, we also need to simultaneously hear the sound of the 'suona' and see it in visual input. At this stage, any temporal misalignment between sound and visual appearance of the 'suona' can hamper learning. Our method successfully learns the Spatial and Temporal levels and then, it also relates both the sound and appearance with the word 'suona' on the Semantic level. In Figure \ref{fig:3}\textbf{v}, there are two instruments i.e. xylophone and piano and the question is about 'appeared' not 'played' as only the xylophone was played and to answer correctly, the model requires Semantic level learning which ST-AVQA \cite{li2022learning} lacks and predicts 'one' which is the wrong answer. These results demonstrate the ability of our method, as shown in quantitative evaluation, to perform better than the state-of-the-art in AV and V scenes and also, performs equally well in the audio (A) category.

\subsection{Ablation Results and Discussion}
\label{ssec:ablation results}
In this section, we discuss the effect of our contributions. This includes the Contextual block, AV fine temporal alignment and the network of three cross-attention modules.

\noindent
\textbf{Contextual Block.} Section \ref{ssec:vd_block} emphasizes the importance of incorporating both audio and visual streams in learning from video data, as they offer complementary information. To achieve better Spatial level learning, a Contextual block is introduced, enabling the network to extract spatio-temporal visual context. The output of the Contextual block is fed to the cross-attention modules, facilitating the unified representation of class features from different modalities. In Table \ref{table:2}, 'w/o Contextual block' clearly demonstrates the effectiveness of the Contextual block. Similarly, 'w ST-AVQA \cite{li2022learning} + Contextual block' shows the effectiveness of the Contextual block when it is added to the existing method ST-AVQA \cite{li2022learning}. It is important to note that while re-implementing ST-AVQA \cite{li2022learning} with own our contributions, we employ training parameters and feature extractors used by our method for a fair comparison.

\noindent
\textbf{Pre-training - AV Fine Temporal Alignment (AVFA).} Section \ref{ssec:mmfa} demonstrates the significance of fine temporal alignment between audio and visual streams. A pre-training task based on AV fine alignment plays an important role in improved performance. In Table \ref{table:2}, entry 'w/o Pre-training' shows the effectiveness of this. Similarly, 'w ST-AVQA \cite{li2022learning} + Pre-training' shows the effectiveness of AVFA when re-implemented with ST-AVQA \cite{li2022learning}.

\noindent
\textbf{Three Cross-Attention Blocks (3CA).} 
The three cross-attention blocks in our architecture play a significant role in Semantic level learning and also, in addressing questions in dynamic scenarios. Two blocks capture information from audio and visual modalities, aligning it with the query to facilitate a comprehensive understanding of the question. The third block propagates audio information to the visual stream, enhancing integration and feature alignment. By incorporating all three modules, we create a robust framework that improves overall performance by covering audio, visual, and AV dynamic scenarios effectively. In Table \ref{table:2}, entry 'w 3CA only' shows the effectiveness of this contribution without the other two contributions. Also, 'w 2CA' shows the limitation of the network with two cross-attention blocks using text as query and audio and visual as key, value for each, while using both AVFA and the Contextual block. Similarly, 'w 4CA' shows decreased performance when using another cross-attention module in addition to 'w 3CA' where visual semantic is employed as a query and audio as both key and value.

\noindent
\textbf{Effect of Input.} 
The last 4 rows of the table show the effect of different input combinations on our proposed method. 'w Q' is when we only send questions (Q) as input. The model uses information available within the question to predict the answers. This is also the manifestation of Figure \ref{fig:1} where the same class but different modalities are closer to each other and even one modality as an input can help in an accurate answer. Rows 'w A + Q' and 'w V + Q' show  results for audio and visual category questions which perform well when given their respective inputs i.e., audio + question (A + Q) and visual + question (V + Q) but perform adversely for the inverse case. The performance is balanced for the audio-visual question category for these two rows. The last row demonstrates performance with all the modalities given as input, demonstrating the best performance.

\section{Conclusion}
We introduce a novel CAD network for AVQA task. We proposed a parameter-free spatial alignment block, temporally aligned pre-training, and semantic audio-visual balance. The CAD network boosts performance on MUSIC-AVQA dataset against state-of-the-art methods, showcasing enhanced robustness and efficiency. Our work improves on the AVQA task and the contributions can find applications in other tasks such as video captioning, speech/speaker recognition, action recognition, etc. Our work addresses the challenge of AV misalignment at Spatial, Temporal and Semantic level, which introduces inaccuracies in the understanding of the AV information, potentially resulting in biased decision-making. These contributions can potentially improve accessibility for individuals with sensory impairments.
\section{Acknowledgement}
This research was partly supported by the British Broadcasting Corporation Research and Development (BBC R\&D), Engineering and Physical Sciences Research Council (EPSRC) Grant EP/V038087/1 “BBC Prosperity Partnership: Future Personalised Object-Based Media Experiences Delivered at Scale Anywhere”.

{\small
\bibliographystyle{ieee_fullname}
\bibliography{egbib}
}

\newpage
\noindent
\title{\textbf{Appendix}}
\\
\\
\noindent
\textbf{Datasets and Implementation Details} 

\textbf{Dataset.} For AV fine temporal alignment-based pre-training, we employ ACAV100M\cite{lee2021acav100m} dataset. The ACAV100M dataset is a very large-scale collection of 100 million audio-visual clips designed for audio-visual representation learning. It covers diverse topics including human sounds, music, animal, and nature sounds, etc and facilitates the exploration of synchronisation, alignment, and semantic associations between audio and visual streams. We leverage this dataset to advance the state-of-the-art in dynamic audio-visual question answering (AVQA), one of the toughest applications in audio-visual learning, by pre-training our novel network on the music category of this dataset. \\

\textbf{Implementation.} The music category in ACAV100M dataset contains around 26.3 million videos and we sample 6 million videos out of these. Each video has a length of 10 seconds and for our pre-training task, we stitch 6 videos together. Then, we create 60 cues of one second each and in the next step, we extract features of audio and visual streams using PANNs \cite{kong2020panns} and ViT \cite{dosovitskiy2020image} respectively. \\

This dataset does not contain any annotations and for question queries, first, we use GIT\textsubscript{L} \cite{wang2022git},  with standard settings, to generate a caption for each 10-second video out of the total 6 videos which are stitched into one. In the next step, we generate a question query as did in BEIR \cite{thakur2021beir} from each caption. In this manner, we end up with 6 question queries for a combined 60-second video where each query belongs to each 10-second clip. We use question queries in a way that when we send a positive pair (audio and visual of the same cue) as input, a question query of that 10-second clip that contains the cue is employed. In the case of the negative pair (audio and visual of the different cues), we alternatively use the question query of either the audio stream clip or the visual stream clip if they belong to different clips. Here, the question query is encoded using \cite{pennington2014glove}. \\

\begin{figure}[h]
  \centering
   \includegraphics[width=\linewidth]{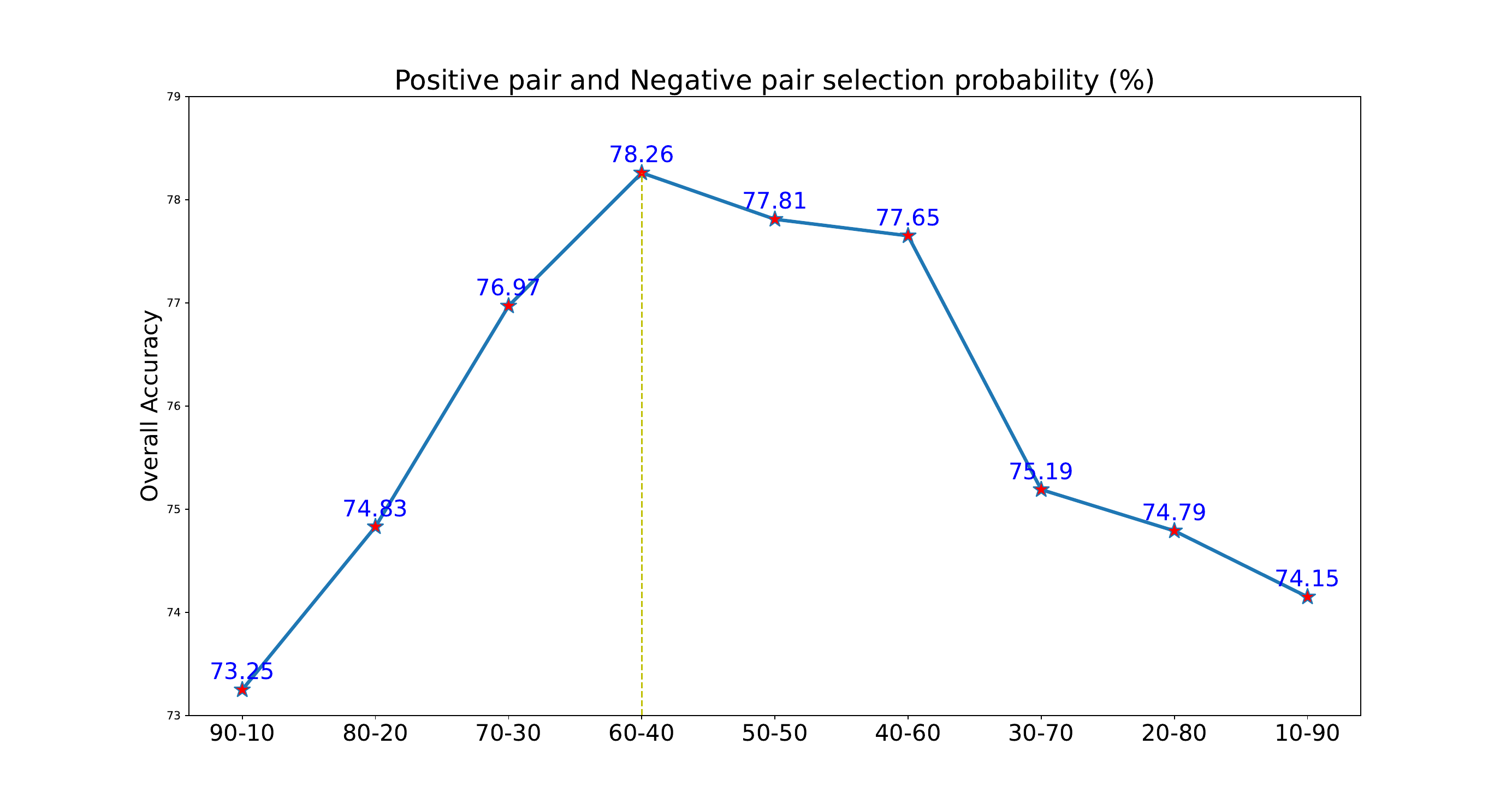}
   \vspace{-5mm}
   \caption{Effect of positive and negative pairs selection probability on the overall performance.}  
   \label{fig:5}\vspace{1mm}
\end{figure}

We use selection probability of 60\% and 40\% (60-40) for the positive and negative pairs respectively. In Figure \ref{fig:1}, we demonstrate this by evaluating different combinations of positive-negative pairs' selection probability in \%. We start with a 90-10 split and pre-train the model till 10-90. For each split configuration, we evaluate it by training the model and then, testing it to find the overall accuracy. The model achieved the highest accuracy for the 60-40 split as shown in Figure \ref{fig:5}.\\

\begin{figure}[h]
  \centering
   \includegraphics[width=\linewidth]{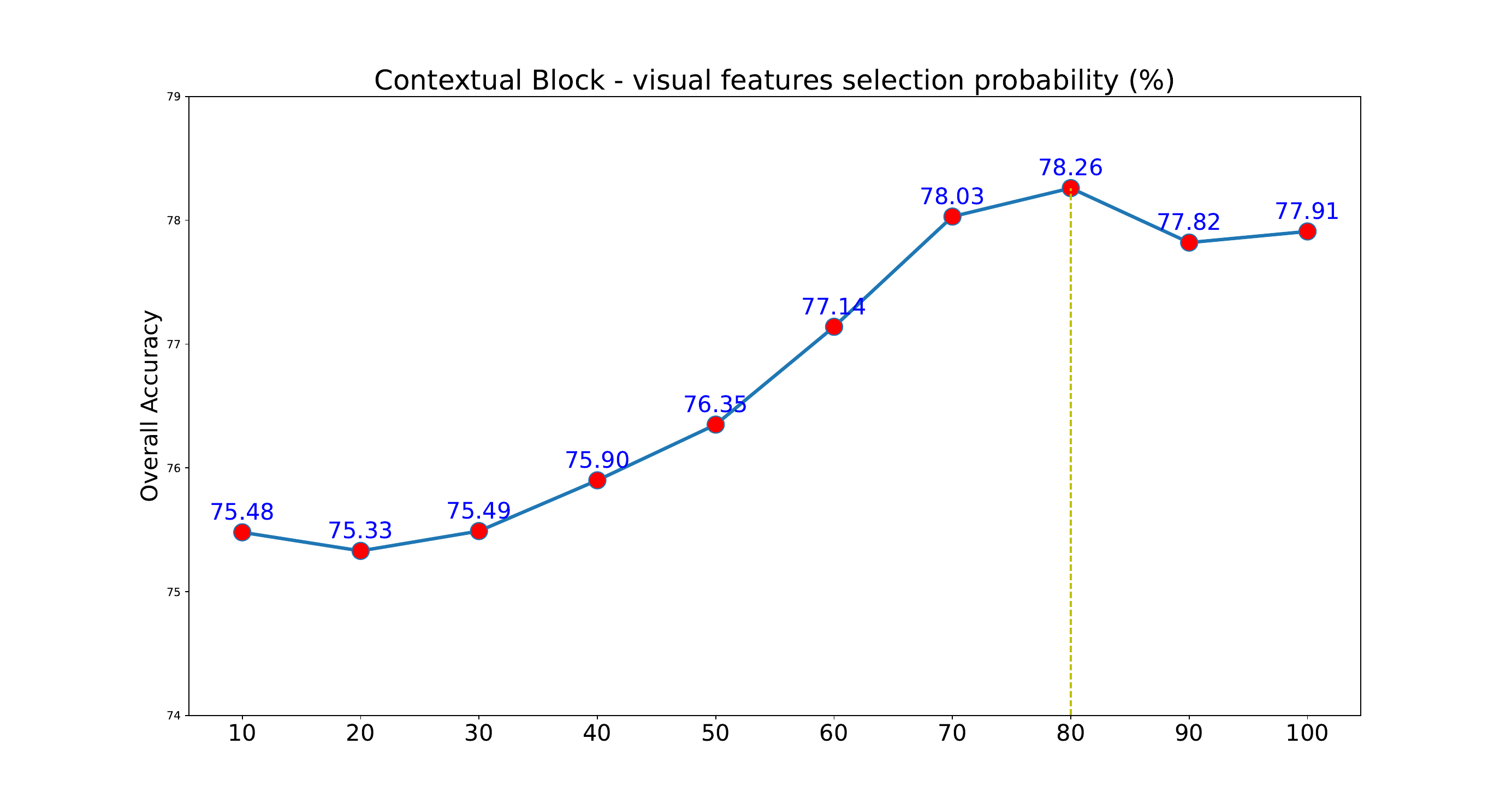}
   \vspace{-5mm}
   \caption{Effect of \% of visual features to go through Contextual Block on the overall performance.}  
   \label{fig:6}\vspace{1mm}
\end{figure}

Another contribution to our work is the Contextual Block where we select 80\% of the visual features to be passed through it. This is based on the experiments as well as shown in Figure \ref{fig:6}. We iterate from 10\% to 100\% where initially there is no effect on the performance but it picks up from 40\% to 80\% and beyond that it decreases slightly.\\

\begin{figure}[h]
  \centering
   \includegraphics[width=\linewidth]{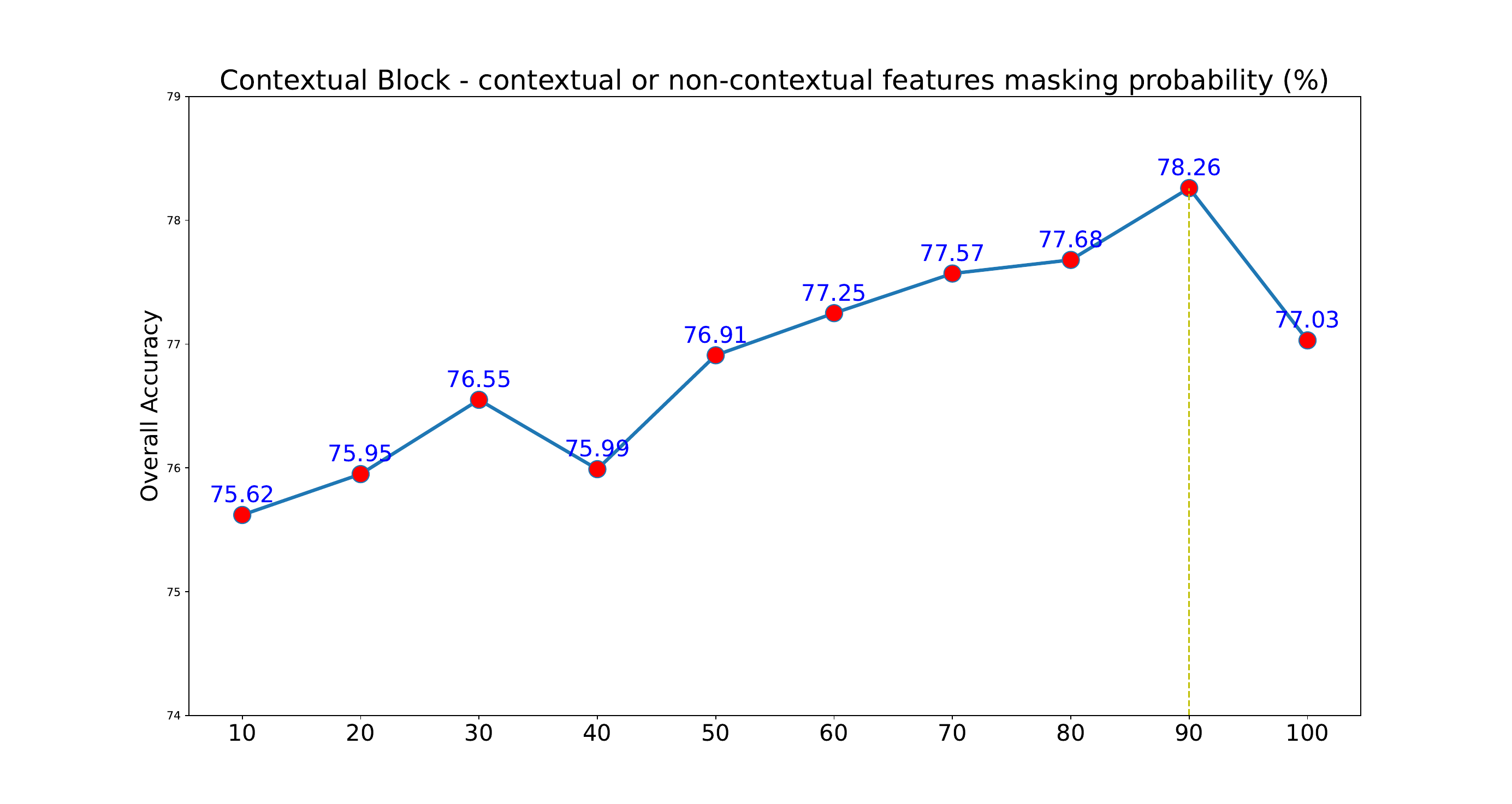}
   \vspace{-5mm}
   \caption{Effect of \% of contextual or non-contextual features masked in the Contextual Block on the overall performance.}  
   \label{fig:7}\vspace{1mm}
\end{figure}

The last of these experiments is the selection of \% of either contextual or non-contextual features for masking within the Contextual Block. In this work, we zero out or mask 90\% based on the experimentation shown in Figure \ref{fig:7}. \\

We also calculate the GPU hours of pre-training and training stages for our method (CAD) as well as the state-of-the-art (SOTA) \cite{li2022learning} to analyse the effect of the Contextual Block. As shown in Table \ref{table:4}, GPU hours decrease for both the pre-training and training stages for our method as well as the SOTA when implemented with the Contextual Block. In this case, we implement the SOTA with the pre-training and also, the Contextual Block.  

\begin{table}[H]
\centering
\resizebox{\linewidth}{!}{%
\begin{tabular}{*{1}c|*{1}c|*{1}c}
\hline
\multicolumn{1}{c|}{Task}  & \multicolumn{1}{c|}{Pre-training (GPU hours)} & \multicolumn{1}{c}{Training (GPU hours)}                                                                                                 \\ \hline 
                        \multicolumn{1}{c|}{w/o Contextual Block - ST-AVQA (SOTA) \cite{li2022learning}}    & \multicolumn{1}{c|}{205.7}  & \multicolumn{1}{c}{2.1}                 \\ \hline 
\multicolumn{1}{c|}{w Contextual Block - ST-AVQA (SOTA) \cite{li2022learning}}          & \multicolumn{1}{c|}{131.2} & \multicolumn{1}{c}{1.6} \\  \hline
\multicolumn{1}{c|}{w/o Contextual Block - CAD (Ours)}          & \multicolumn{1}{c|}{257.1} & \multicolumn{1}{c}{2.4} \\  \hline
\multicolumn{1}{c|}{w Contextual Block - CAD (Ours)}          & \multicolumn{1}{c|}{149.4} & \multicolumn{1}{c}{1.7} \\ 
\end{tabular}%
}\vspace{2mm}
\parbox[t]{\linewidth}{\caption{w/o Contextual Block and w Contextual Block describe the effect of without and with the Contextual Block on the employed compute for the SOTA \cite{li2022learning} as well as our method CAD.} 
\label{table:4}}
\end{table}

By comparing the first and last row of Table \ref{table:4}, our method demonstrates efficiency over the SOTA. When coupled with the Contextual Block, the SOTA also demonstrates more efficiency as shown in row 2 but the performance is still lower than our method as demonstrated in Table 2 of the main paper.

\end{document}